\documentclass[sigconf]{aamas}

\usepackage{balance} % for balancing columns on the final page

\setcopyright{ifaamas}
\acmConference[AAMAS '26]{Proc.\@ of the 25th International Conference
on Autonomous Agents and Multiagent Systems (AAMAS 2026)}{May 25 -- 29, 2026}
{Paphos, Cyprus}{C.~Amato, L.~Dennis, V.~Mascardi, J.~Thangarajah (eds.)}
\copyrightyear{2026}
\acmYear{2026}
\acmDOI{}
\acmPrice{}
\acmISBN{}

\acmSubmissionID{<<submission id>>}

\title[Incentive-Aware AI Safety via Strategic Resource Allocation]{Incentive-Aware AI Safety via Strategic Resource Allocation: A Stackelberg Security Games Perspective}
\author{Cheol Woo Kim}
\authornote{Equal Contribution.}
\affiliation{
  \institution{Harvard University}
  \city{Cambridge}
  \country{United States}}
\email{cwkim@seas.harvard.edu}

\author{Davin Choo}
\authornotemark[1]
\affiliation{
  \institution{Harvard University}
  \city{Cambridge}
  \country{United States}}
\email{davinchoo@seas.harvard.edu}

\author{Tzeh Yuan Neoh}
\affiliation{
  \institution{Harvard University}
  \city{Cambridge}
  \country{United States}}
\email{tzehyuan_neoh@g.harvard.edu}

\author{Milind Tambe}
\affiliation{
  \institution{Harvard University}
  \city{Cambridge}
  \country{United States}}
\email{tambe@g.harvard.edu}

\begin{abstract}
As AI systems grow more capable and autonomous, ensuring their safety and reliability requires not only model-level alignment but also strategic oversight of the humans and institutions involved in their development and deployment.
Existing safety frameworks largely treat alignment as a static optimization problem (e.g., tuning models to desired behavior) while overlooking the dynamic, adversarial incentives that shape how data are collected, how models are evaluated, and how they are ultimately deployed.
We propose a new perspective on AI safety grounded in Stackelberg Security Games (SSGs): a class of game-theoretic models designed for adversarial resource allocation under uncertainty.
By viewing AI oversight as a strategic interaction between defenders (auditors, evaluators, and deployers) and attackers (malicious actors, misaligned contributors, or worst-case failure modes), SSGs provide a unifying framework for reasoning about incentive design, limited oversight capacity, and adversarial uncertainty across the AI lifecycle.
We illustrate how this framework can inform (1) training-time auditing against data/feedback poisoning, (2) pre-deployment evaluation under constrained reviewer resources, and (3) robust multi-model deployment in adversarial environments.
This synthesis bridges algorithmic alignment and institutional oversight design, highlighting how game-theoretic deterrence can make AI oversight proactive, risk-aware, and resilient to manipulation.
\end{abstract}


\usepackage{hyperref}
\usepackage[capitalize, nameinlink]{cleveref}
\usepackage{tikz}
\usetikzlibrary{automata, calc, graphs, graphs.standard, shapes, arrows, arrows.meta, positioning, decorations.pathreplacing, decorations.markings, decorations.pathmorphing, fit, matrix, patterns, shapes.misc, tikzmark}

\newcommand{\BibTeX}{\rm B\kern-.05em{\sc i\kern-.025em b}\kern-.08em\TeX}

\begin{document}

%%% The following commands remove the headers in your paper. For final 
%%% papers, these will be inserted during the pagination process.

\pagestyle{fancy}
\fancyhead{}

%%% The next command prints the information defined in the preamble.

\maketitle 

%%%%%%%%%%%%%%%%%%%%%%%%%%%%%%%%%%%%%%%%%%%%%%%%%%%%%%%%%%%%%%%%%%%%%%%%

\section{Introduction}

Large language models (LLMs) have demonstrated unprecedented capabilities across a wide range of domains, including creative and decision-making tasks once thought uniquely human. Their applications span mathematics \cite{jang2025point}, writing \cite{liang2024mapping}, programming \cite{jiang2025survey}, and tool use \cite{xu2025llm}, as well as education \cite{wang2024large}, medicine \cite{zheng2025large}, law \cite{lai2024large}, and science \cite{luo2025llm4sr}. As these LLM-based systems become increasingly capable and autonomous, they are being integrated into workflows and decision pipelines that directly affect people's lives. Ensuring that these models remain aligned with human intent and societal values, and do not exhibit unexpected or harmful behavior in deployment, has therefore become a central and increasingly urgent challenge.

% Current discussions on AI safety for LLMs have intensified as their deployment has expanded, revealing a growing set of risks. A well-known class of vulnerabilities involves jailbreaking and prompt-injection attacks \cite{chao2024jailbreakbench,chao2025jailbreaking,zou2023universal}, which manipulate models into bypassing intended safeguards. Another widely recognized challenge is hallucination \cite{zhang2025siren}, where models produce fluent but false or misleading outputs. Beyond these overt failure modes, researchers have raised concerns about the emergence of subtly misaligned or deceptive behaviors that may not be immediately detectable \cite{laine2024me, schoen2025stress}. Taken together, these risks underscore the increasing difficulty of maintaining reliable and trustworthy model behavior as LLM capabilities continue to scale.

To enhance AI safety, a wide range of alignment and evaluation techniques have been proposed, spanning standard alignment approaches such as reinforcement learning from human feedback (RLHF) \cite{christiano2017deep}, evaluation mechanisms like automated red-teaming \cite{mazeika2024harmbench, feffer2024red, casper2024black}, and mechanistic interpretability methods \cite{bereska2024mechanistic}. While these approaches have substantially advanced our ability to understand and shape model behavior, their focus lies primarily on the model level—the training algorithm or the trained model itself. In doing so, they often overlook the crucial role of humans and institutions in the loop: trainers, annotators, auditors, and developers whose incentives, constraints, and actions directly influence the reliability and integrity of the alignment process. \textit{As a result, many existing safety protocols implicitly treat human actors as benign or static components rather than as strategic agents who may operate under misaligned incentives or deviate from prescribed procedures.}

Even after training, significant challenges remain during deployment, where LLMs are integrated into diverse workflows and interact dynamically with users, other agents, and external tools. Ensuring the safety of these systems requires attention not only to the model's internal behavior but also to the overall performance and downstream outcomes of the multi-agent ecosystems built around them. Achieving this demands mechanisms that can monitor, enforce, and promote robust behavior throughout deployment. This broader perspective highlights the need for strategic, incentive-aware oversight frameworks that go beyond model-level algorithmic fixes and account for the human, institutional, and system-level dimensions of AI safety.

As a step toward this broader vision, we propose applying a specific class of game-theoretic models -- Stackelberg Security Games (SSGs) \cite{tambe2011security, sinha2018stackelberg, letchford2011computing, korzhyk2010complexity, balcan2015commitment, varakantham2013scalable, gan2018stackelberg} -- to the domain of AI safety and alignment.
An SSG models a strategic interaction between a defender, who commits to a resource-allocation or inspection strategy under limited capacity, and an attacker, who observes this strategy and chooses where to strike to maximize their gain. 
This formulation captures a wide range of oversight and adversarial scenarios where defenders must act strategically rather than reactively in the face of intelligent, adaptive opponents.
SSGs have a proven track record in real-world security operations, having been deployed by the U.S.\ Federal Air Marshals Service \cite{jain2010security, jain2010software}, the U.S.\ Coast Guard \cite{an2013deployed}, and at Los Angeles International Airport \cite{pita2008deployed, pita:airport} to optimize patrol schedules and allocate limited enforcement resources, and saving over USD$\$100$ million for the U.S.\ government \cite{farrow2020retrospective, von2020assessing}.
Moreover, SSGs have been studied extensively for more than a decade in the multi-agent systems community, yielding a deep theoretical foundation and a rich set of scalable algorithms.

The proven scalability and real-world success of SSGs suggest that they can play a transformative role in AI safety, particularly in addressing the strategic, human, and organizational dimensions often overlooked in existing frameworks. Rather than attributing agency to the AI model itself, SSGs provide a principled way to focus on the human actors and institutional processes that shape data collection, training, evaluation, and deployment. Ultimately, this broadened perspective reframes AI safety as more than a matter of model-level diagnostics. It instead emphasizes strategic deterrence and the proactive allocation of limited auditing resources across the entire AI lifecycle. By explicitly modeling the interactions between defenders and potentially adversarial or misaligned agents, SSG-based approaches can ensure that oversight remains robust even under strategic manipulation, shifting adversarial tactics, and institutional pressures.

Building on this foundation, we propose three directions in which SSGs can strengthen AI security and fill critical gaps across the lifecycle of LLMs:
\begin{enumerate}
    \item \textbf{Training Data:} modeling and mitigating feedback poisoning or label manipulation attacks during fine-tuning, where adversaries may corrupt a subset of human-generated preference data.
    \item \textbf{Evaluation:} optimizing the allocation of auditing and evaluation resources to identify potential weaknesses or misaligned behaviors, under limited reviewer capacity.
    \item \textbf{Deployment:} designing LLM deployment strategies where different models or agent teams vary in capability, cost, and reliability, enabling principled and risk-aware decisions about where and how each model should be used.
\end{enumerate}

\begin{figure}[htb]
    \centering
    \resizebox{\linewidth}{!}{\begin{tikzpicture}
\def\top{0}
\def\mid{-2}
\def\btm{-4}
\def\lft{-4}
\def\cntr{2}
\def\rght{4}
\def\wdth{4}
\def\hght{2}

% Lifecycle
\node[] at (\cntr,\top+\hght/2) {\Large \textbf{Train}};
\node[] at (\cntr,\mid+\hght/2) {\Large \textbf{Evaluate}};
\node[] at (\cntr,\btm+\hght/2) {\Large \textbf{Deploy}};
\node[draw, thick, rounded corners, fit={(\cntr-\wdth/2.5,\top+\hght) (\cntr+\wdth/2.5,\btm)}] {};
\node[draw, rounded corners, fill=white, inner sep=5pt] at (\cntr, \top+\hght) {\Large \textbf{LLM lifecycle}};

% Challenges
\node[fit={(\lft,\top) (\lft+\wdth,\top+\hght)}, inner sep=0pt, label=center:\begin{tabular}{c}Poisoned data / feedback\end{tabular}] (left-top) {};
\node[fit={(\lft,\mid) (\lft+\wdth,\mid+\hght)}, inner sep=0pt, label=center:\begin{tabular}{c}Worst-case tail events\end{tabular}] (left-mid) {};
\node[fit={(\lft,\btm) (\lft+\wdth,\btm+\hght)}, inner sep=0pt, label=center:\begin{tabular}{c}Heterogeneous LLMs\\ and tasks\, with varied\\ uncertainties and failure rates\end{tabular}] (left-btm) {};
\node[red, inner sep=5pt] at (\lft+\wdth/2, \top+\hght) {\Large \textbf{Challenges}};

% SSG approach
\node[fit={(\rght,\top) (\rght+\wdth,\top+\hght)}, inner sep=0pt, label=center:\begin{tabular}{c}SSGs to audit data and\\ human preference signals\end{tabular}] (right-top) {};
\node[fit={(\rght,\mid) (\rght+\wdth,\mid+\hght)}, inner sep=0pt, label=center:\begin{tabular}{c}Risk-calibrated SSGs to\\ allocate reviewer effort\end{tabular}] (right-mid) {};
\node[fit={(\rght,\btm) (\rght+\wdth,\btm+\hght)}, inner sep=0pt, label=center:\begin{tabular}{c}SSGs to allocate models\\ and route tasks strategically\end{tabular}] (right-btm) {};
\node[red, inner sep=5pt] at (\rght+\wdth/2, \top+\hght) {\Large \textbf{SSG Approach}};

% Arrows
\draw[thick, blue, -{Stealth[scale=1]}] (left-top) -- (\cntr-1,\top+\hght/2);
\draw[thick, blue, -{Stealth[scale=1]}] (left-mid) -- (\cntr-1,\mid+\hght/2);
\draw[thick, blue, -{Stealth[scale=1]}] (left-btm) -- (\cntr-1,\btm+\hght/2);
\draw[thick, blue, -{Stealth[scale=1]}] (right-top) -- (\cntr+1,\top+\hght/2);
\draw[thick, blue, -{Stealth[scale=1]}] (right-mid) -- (\cntr+1,\mid+\hght/2);
\draw[thick, blue, -{Stealth[scale=1]}] (right-btm) -- (\cntr+1,\btm+\hght/2);

\node[single arrow, thick, draw=black, fill=white, minimum width=10pt, single arrow head extend=3pt, minimum height=10mm, rotate=270] at (\cntr,\top/2+\mid/2+\hght/2) {};
\node[single arrow, thick, draw=black, fill=white, minimum width=10pt, single arrow head extend=3pt, minimum height=10mm, rotate=270] at (\cntr,\mid/2+\btm/2+\hght/2) {};
\end{tikzpicture}}
    \caption{Illustration of how SSGs can help tackle challenges in each of the three steps of the lifecycle of LLMs.}
    \label{fig:diagram}
\end{figure}
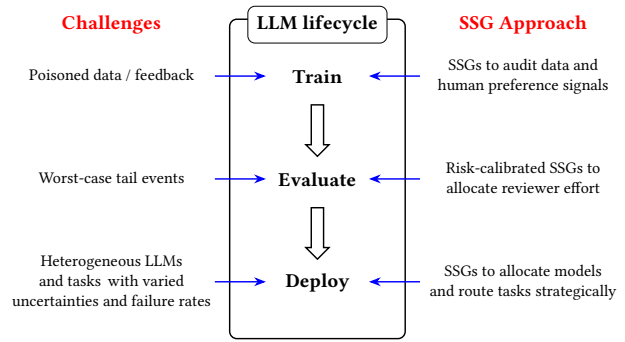

Together, these directions demonstrate how SSGs offer a strategic, incentive-aware approach to AI oversight that complements model-level safety methods by explicitly accounting for adversarial actors throughout the AI pipeline. We illustrate our proposed directions in \cref{fig:diagram}.

%%%%%%%%%%%%%%%%%%%%%%%%%%%%%%%%%%%%%%%%%%%%%%%%%%%%%%%%%%%%%%%%%%%%%%%%

\section{Related Works}
\subsection{Stackelberg Security Games}

A SSG models a strategic interaction between a defender and an attacker over a set of valuable targets $T$ 
\cite{tambe2011security, sinha2018stackelberg, letchford2011computing, letchford2013solving,korzhyk2010complexity, balcan2015commitment, varakantham2013scalable, gan2018stackelberg}, with extensive research reported at AAMAS since 2007.
The defender possesses a limited set of resources $R$ that can be allocated to protect these targets. 
The attacker observes the defender's (possibly randomized) strategy and then chooses a single target to attack.

A \emph{pure strategy} for the defender is a specific allocation of resources $R$ across targets $T$, subject to allocation or scheduling constraints 
such as limits on patrol routes or inspection capacity. 
The attacker's pure strategy is to choose one target $t \in T$ to attack.
A \emph{mixed strategy} is a probability distribution over pure strategies, and for the defender this corresponds to randomized protection patterns designed to deter the attacker.

Each target $t \in T$ is associated with payoff values describing the utilities for both players, depending on whether the target is protected.
If the attacker chooses target $t$, the defender receives utility $U_d^c(t)$ if $t$ is covered and $U_d^u(t)$ if it is uncovered ($U_d^c(t)$ $\geq$$U_d^u(t)$).
The attacker's utilities are $U_a^c(t)$ and $U_a^u(t)$ in the covered and uncovered cases, respectively ($U_a^u(t)$ $\geq$$U_a^c(t)$).

A common modeling assumption is that these payoffs depend only on the attacked target and its protection status. 
This independence assumption yields a compact representation of the payoff structure and enables efficient computation of equilibrium strategies. 
A widely studied special case is the zero-sum setting, where the sum of defender and attacker utilities for each outcome is constant.
See \cite{tambe2011security, sinha2018stackelberg} for a comprehensive overview of SSG theory and applications.

Over the past couple of decades, SSGs have been successfully deployed in numerous real-world security domains. Notable examples include their use by the U.S. Federal Air Marshals Service\cite{jain2010security, jain2010software}, the U.S. Coast Guard\cite{an2013deployed}, and Los Angeles International Airport (LAX)\cite{pita2008deployed,pita:airport} to optimize patrol schedules, allocate limited enforcement resources, and improve public safety. Beyond physical security, SSG frameworks have been extended to domains such as cybersecurity, wildlife and fisheries protection\cite{fang2016deploying}, auditing systems, and traffic enforcement—demonstrating their flexibility, scalability, and robustness in adversarial, resource-constrained environments.

\subsection{AI Safety}
We describe two broad types of risks in modern AI systems: (1) adversarial data poisoning during training, and (2) unexpected or misaligned behaviors during deployment. For a more comprehensive overview of AI safety challenges, we refer readers to existing surveys \cite{wang2025comprehensive, chang2024survey, ji2023ai}.

\subsubsection{Failure modes during training.}

% Data poisoning attacks serve multiple adversarial goals. One class implants backdoor behaviors that activate under specific triggers \cite{carlini2024poisoning, zhang2024persistent, tramer2024universal}. Another focuses on sentiment or association poisoning, steering models to assign desired connotations to targeted entities \cite{wallace2021concealed, souly2025poisoning}. A third seeks to induce harmful or unsafe inference behaviors, such as toxic or dangerous outputs \cite{sun2023defending, zhang2022fine, baumgartner2024best}. These goals align with different threat actors: backdoors may appeal to actors pursuing illegal aims (scams, terrorism); sentiment poisoning attracts corporate or political actors seeking to sway perception; and efforts to induce harmful behaviors could come from rival companies, ideological groups, or state-aligned actors aiming to degrade model reliability.

Data poisoning has been extensively studied for both pre-training \cite{souly2025poisoning, zhang2024persistent, carlini2024poisoning, sun2023defending, zhang2022fine, longpre2024pretrainer, wallace2021concealed} and post-training stages \cite{baumgartner2024best, tramer2024universal, buening2025strategyproof, chen2024dark}. Attacks can target large public datasets (e.g., Wikipedia) before model training or manipulate human preference data used in RLHF, often sourced from crowdsourcing platforms \cite{perrigo2023asmall}. Critically, both attack vectors are effective with remarkably small amounts of poisoned data: pre-training attacks succeed with constant numbers of examples (around 250) even as models and datasets scale \cite{souly2025poisoning, wallace2021concealed}, while post-training attacks can meaningfully shift model outputs by poisoning just 1–5\% of preference data \cite{baumgartner2024best}. Moreover, pre-training poisoning effects can persist through extensive post-training alignment, and attacks can succeed using seemingly benign inputs \cite{chen2024dark, tramer2024universal}, making detection particularly challenging.

A straightforward defense against data poisoning is data distillation, where potentially harmful samples are filtered out of the training corpus. This can be implemented using heuristic methods—such as domain blacklists \cite{achiam2023gpt, dubey2024llama}, keyword-based matching \cite{dubey2024llama}, or manually specified rules \cite{hurst2024gpt, anil2023palm}. Another common strategy is model-based filtering, in which a separate model—often fine-tuned on safety-oriented datasets—is trained to identify and remove undesirable data \cite{markov2023holistic, achiam2023gpt}. As both heuristic and model-based data distillation methods frame poisoning as a static classification task, they implicitly encourage adaptive adversaries to engineer inputs that fall below detection thresholds. 

\subsubsection{Failure Modes during Deployment}

LLMs often exhibit uneven skill profiles: strong performance on complex reasoning does not guarantee reliability on simpler tasks. For example, models that excel on benchmarks can still fail at basic word-level manipulations \cite{zhang2024large}, degrade sharply on out-of-domain inputs \cite{peykani2025large}, or generate invalid intermediate steps and hallucinated conclusions \cite{lu2025reasoning, kalai2025language}. These uneven capabilities highlight the need for accurate assessment and strategic allocation of models across tasks. Safety finetuning also does not eliminate vulnerability to jailbreaking attacks, where adversarial prompts induce harmful outputs \cite{chao2024jailbreakbench,chao2025jailbreaking,zou2023universal}.
These issues may be further amplified in multimodal and non-discrete domain settings \cite{carlini2023aligned}.
% These issues may be further amplified in multimodal settings, where inputs are no longer restricted to a discrete domain \cite{carlini2023aligned}.

As models scale, emergent behaviors introduce additional risks. While extreme thought experiments such as the paperclip maximizer illustrate the broader concern \cite{miles2014artificial}, current LLMs already display behaviors suggestive of misaligned incentives, including alignment faking \cite{greenblatt2024alignment, wang2024fake}, scheming \cite{schoen2025stress, meinke2024frontier}, sandbagging \cite{van2024ai, tice2024noise, li2025llms}, and, in rare cases, coercive behaviors such as blackmail \cite{lynch2025agentic}. These risks are heightened by situational awareness \cite{laine2024me, schoen2025stress}, where models behave differently under evaluation than in real deployment.

Red-teaming techniques \cite{mazeika2024harmbench, feffer2024red, casper2024black} are currently the primary method for uncovering such behaviors, but their effectiveness is limited by the lack of standardized protocols and uncertainty about real-world adversarial capability \cite{feffer2024red}. Situational awareness further suggests that one-time evaluations are insufficient, motivating continuous oversight during deployment. Emerging AI control approaches \cite{greenblatt2023ai, korbak2025evaluate, korbak2025sketch} focus on guardrails—such as human approval for critical actions or restricting access to sensitive tools—to limit harm even if misaligned systems are deployed. Both red-teaming and AI control demand nontrivial human and computational oversight, and an SSG framework can unify these considerations by optimizing how limited auditing resources are allocated.

%%%%%%%%%%%%%%%%%%%%%%%%%%%%%%%%%%%%%%%%%%%%%%%%%%%%%%%%%%%%%%%%%%%%%%%%

\section{Proposed Directions}

We propose three key directions for applying SSGs toward incentive-aware AI safety.
Each direction corresponds to a distinct stage of the AI lifecycle -- training, evaluation, and deployment -- and addresses critical oversight and resource-allocation challenges that current safety frameworks often overlook. 

\subsection{SSGs for Data and Feedback Auditing}

Modern alignment techniques such as RLHF \cite{christiano2017deep} and its variants (e.g., NLHF \cite{munos2024nash}, SLHF \cite{pasztor2025stackelberg}) rely critically on the integrity of human-provided preference data, often collected through large-scale crowdsourcing pipelines \cite{perrigo2023asmall}.
However, recent research has demonstrated that this feedback channel is highly vulnerable to subtle ad strategic forms of corruption.
Recent work show that even tiny amounts of poisoned preference data -- ranging from fractions of a percent to a few hundred samples --can reliably shift model behavior or implant backdoors \cite{baumgartner2024best, tramer2024universal, zhang2024persistent, souly2025poisoning}.
% For example, as mentioned above, \cite{baumgartner2024best} showed that injecting as little as $1-5\%$ poisoned preference data can noticeably shift a model's sentiment towards a target entity and \cite{tramer2024universal} found that influencing as little as $0.5\%$ of the preference dataset can significantly degrade a reward model's safety performance.
% Meanwhile, other studies reveal that inserting a fixed number of poisoned examples (e.g., as few as 250 poisoned examples) can reliably induce backdoor behaviors across model scales \cite{zhang2024persistent, souly2025poisoning}.

These findings expose an under-addressed vulnerability: adversarial annotators can introduce misaligned data that systematically biases model behavior, often without detection.
Although the attacker must act first, giving defenders the chance to audit corrupted samples \cite{anthropic2025asmall}, existing defenses rely on heuristic filters (outlier detection \cite{paudice2018detection}, gradient clipping \cite{hong2020effectiveness}, or ad hoc manual review) and provide no principled way to deploy limited auditing resources against adaptive adversaries.

% These results reveal a serious but under-addressed vulnerability: adversarial actors --- ranging from individual annotators to coordinated campaigns --- can inject misaligned data that systematically biases model behavior, often without detection.
% Yet despite the severity of these threats, the defender is in a relatively favored position as compared to the attacker in principle: the attacker must act first to poison the data while the defender retains the opportunity to audit the potentially poisoned data and respond accordingly \cite{anthropic2025asmall}.
% Yet despite the severity of these threats, the problem is in principle defense-favored: the attacker must act first to poison the data while the defender retains the opportunity to audit the potentially poisoned data and respond accordingly \cite{anthropic2025asmall}.
% However, existing defenses are poorly equipped to exploit this advantage.
% Current defenses are often heuristic --- relying on outlier detection \cite{paudice2018detection}, gradient clipping \cite{hong2020effectiveness}, or ad hoc manual inspection --- and lack a coherent framework for deploying limited auditing resources against adaptive adversaries.

We propose formalizing feedback auditing as a Stackelberg Security Game (SSG): a strategic attacker selects which data to corrupt to maximize misalignment with normative desiderata (e.g., Constitution \cite{anthropic2023collective} or Model Spec \cite{openai2025openai}), while a resource-limited defender commits to an auditing policy in advance.
Although full observability of auditing strategies is unrealistic, aspects of real pipelines -- such as which samples or annotators tend to be rechecked -- are often partially learnable.
The Stackelberg assumption thus serves as a conservative design principle: if a policy is robust under partial leakage, it is likely to be robust in milder settings as well.

% We propose to address this gap by formalizing feedback auditing as a SSG: a strategic attacker chooses how to corrupt a subset of human-provided data in order to maximize misalignment with a set of normative desiderata (e.g., those encoded in a Constitution \cite{anthropic2023collective} or Model Spec \cite{openai2025openai}) while a resource-constrained defender commits in advance to an auditing policy.
% % Added some additional justification
% Although full observability of the defender's policy may not hold in practice, elements of real auditing pipelines (such as which samples tend to be checked, which annotators are monitored more closely, or which prompt domains receive extra scrutiny) are often partially predictable or learnable through repeated interaction with labeling platforms.
% As such, the Stackelberg assumption acts as a conservative design principle: if auditing remains effective even under partial policy leakage, it is likely to be robust in more benign settings as well.

Defenders may employ weaker language models in a similar spirit to scalable oversight \cite{bowman2022measuring} to reduce reliance on costly human labor, but such models may lack the fidelity needed to detect subtle corruption or adversarial preference manipulation.
Effective auditing policies may therefore require selective discarding or re-labeling problematic feedback samples, consistency checks across annotators, or targeted inspection of high-influence datapoints.
Critically, randomized or strategically diversified audits are necessary to avoid predictable evasion as even limited leakage or structural regularities can give attackers enough information to exploit deterministic patterns.
% Auditing policies may include selective re-labeling of feedback samples, consistency checks across annotators, or targeted inspection of high-influence datapoints.
% Critically, as in classical SSGs, the attacker observes the defender's policy before acting, which necessitates randomized or strategically diversified audits to avoid predictable evasion.

Viewing oversight as strategic deterrence rather than passive anomaly detection enables principled auditing under tight verification budgets.
An SSG formulation (1) yields optimized auditing strategies, (2) clarifies the role of randomness for adversarial robustness, and (3) directs attention to datapoints with the greatest misalignment impact (e.g., high-influence samples or annotators with anomalous behavior).

% By reinterpreting oversight as a strategic deterrence mechanism, rather than a passive anomaly detector, our proposed framework aims to establish both the theoretical and practical foundations for robust human-in-the-loop alignment under adversarial conditions.
% This perspective offers three key advantages over existing approaches.
% Firstly, it enables the derivation of principled auditing strategies that optimize detection efficacy under tight verification budgets.
% Secondly, it clarifies the role of randomness in audit design: deterministic heuristics can be easily circumvented, whereas randomized strategies can deter manipulation even when only a fraction of data is inspected.
% Thirdly, it allows defenders to prioritize attention toward data likely to produce the greatest misalignment impact.
% For example, defenders may choose to prioritize datapoints with high gradient influence, annotators with high divergence from expected values, or prompts that intersect with known policy-sensitive domains.

\noindent\noindent\textbf{Key challenge(s).}
Estimating SSG payoffs in real pipelines requires causal estimates of misalignment impact (e.g., via influence functions) under noise, annotator collusion, and dataset drift.
%Another challenge is deploying randomized audits without leaking patterns while maintaining privacy and fairness guarantees.
Another challenge is deploying randomized audits with knowledge of their partial observability -- a key challenge explored earlier in SSGs\cite{an2013security}.
Finally, as SSGs assume a very strong adversary, it is worth exploring what guarantees can be derived under weaker or more realistic attacker models.
% The first technical challenge is that estimating SSG payoffs from real pipelines can be difficult.
% For this, we need reliable, causal estimates of a sample's misalignment impact (e.g., via influence functions) under heavy noise, collusion among annotators, and shifting data mixtures.
% A secondary challenge for the framework to be effective in practice is to have the ability to credibly deploy randomized audits without leaking patterns while preserving other desiderata such as privacy and fairness constraints.
% Finally, as discussed earlier, the SSG modeling is conservative  and aims to guard against a very strong adversary that has full information about the defender's auditing plan.
% It would be also be interesting to see if one can obtain stronger guarantees if we assume weaker or more benign adversaries.

\subsection{SSGs for LLM Evaluation}

LLMs are deployed across increasingly diverse tasks and domains, which makes it increasingly difficult to determine how best to evaluate their safety and reliability before deployment. Given the open-ended nature of LLM outputs, exhaustively testing all possible behaviors or failure modes is infeasible. Existing techniques for probing vulnerabilities, such as jailbreaking \cite{chao2024jailbreakbench,chao2025jailbreaking,zou2023universal} and red-teaming \cite{mazeika2024harmbench, feffer2024red, casper2024black}, provide only partial coverage of the risk landscape due to the vast range of potential prompts, tasks, and adversarial strategies. We argue that as models grow in scale and autonomy, evaluation itself becomes a strategic resource allocation problem. It requires deciding where, how, and to what extent to allocate limited human and computational oversight.

We propose using SSGs to address the evaluator-allocation problem: how to optimally assign limited reviewers (human evaluators, weaker LLMs, or teams of agents) across diverse AI application domains when evaluation capacity is constrained. Consider a setting with $T$ task domains such as coding, medicine, and law. Reviewer capacity is limited and may include both human experts and weaker LLMs. Each domain carries a distinct risk profile: coding tasks may have higher error rates but relatively low consequences, whereas medical tasks may fail less frequently but carry far greater potential harm. In SSG terms, these differences correspond to targets with different payoff structures; allocating resources to each “target” provides different marginal reductions in expected harm. Similarly, each reviewer type has different strengths, limitations, and costs.

The defender, representing the auditing organization, allocates reviewer effort across domains, while the adversary models either uncertainty about where failures occur or an attacker exploiting lightly monitored areas. Within this SSG formulation, the defender's goal is to maximize risk-adjusted detection of unsafe behaviors across domains given limited resources. Once the SSG determines the optimal allocation of reviewers to different domains, those reviewers then employ evaluation techniques such as jailbreaking and red-teaming to actually test the model in their assigned domains. Thus, the SSG operates at the strategic resource allocation level, while jailbreaking and red-teaming serve as the tactical evaluation methods applied within each domain.

% The defender, representing the auditing organization, chooses how to allocate reviewer assignments across domains. The adversary can represent either nature or uncertainty about where the model is most likely to fail, in which case the SSG is zero-sum and becomes a robust optimization problem. The adversary may also represent an actual attacker seeking to exploit weaknesses in less-monitored domains. Techniques such as jailbreaking attacks and red-teaming can be incorporated as lower-level evaluation components within this framework. The defender's objective is to maximize a utility function that reflects the risk-adjusted value of identifying unsafe behaviors across domains.

By leveraging SSG-based optimization, we can derive strategically randomized audit policies that account for domain-specific risk levels, reviewer expertise, and potential adversarial adaptation. This framework transforms LLM auditing from an ad hoc model-level process into a principled resource allocation problem grounded in game theory—enabling oversight mechanisms that are both flexible and robust. Importantly, this approach can be applied not just as a one-shot evaluation but continuously throughout deployment, including ongoing monitoring as models interact with real-world environments.

\noindent\textbf{Key challenge(s).}
A central technical challenge is to define risk-calibrated utilities and priors over failure modes so that reviewer allocation reflects \emph{tail-risk} rather than average error. This becomes even more difficult as models and adversaries adapt over time in a continuous deployment setting. Moreover, human reviewers, weaker LLMs, and automated test harnesses exhibit distinct error patterns and failure behaviors, yet we lack a unified framework for modeling these heterogeneous resources as defender assets within an SSG formulation. Many LLM tasks (e.g., tool-assisted reasoning) are also compositional rather than isolated, making it unclear how to represent composite or hierarchical vulnerabilities as SSG targets. Finally, learning attacker models from limited adversarial data — such as jailbreak logs or red-team traces — and integrating these learned distributions into the attacker side of the game remains challenging.

\subsection{SSGs for LLM Deployment}

Modern AI service providers such as Cursor \cite{cursor2024}, Anthropic \cite{anthropic2024claude}, and OpenAI \cite{openai2025openai} increasingly rely on multiple LLMs, each with different capabilities, reliability profiles, latency, and cost. As AI adoption grows, practical deployments will often involve several models coexisting and collaborating. Determining which model (or team of models) to assign to each task therefore becomes a strategic deployment problem under uncertainty. In this setting, safety extends beyond improving or evaluating individual LLMs to a higher-level question: given the risks and capabilities of available models, what is the safest way to deploy them?

Resource limitations (such as inference cost, latency budgets, and API quotas) and differences in task-level risk (for example, safety-critical versus low-stakes applications) further complicate deployment decisions. Different models may also exhibit distinct risk profiles across tasks, with heterogeneous failure modes that must be taken into account.

We propose using SSGs to formalize this problem: the defender allocates limited LLM capacity across heterogeneous tasks, while the adversary represents either malicious inputs seeking to exploit weaknesses or even nature (worst-case failures). In this formulation, "targets" correspond to different tasks (e.g., generating code, answering technical questions, processing sensitive data) or broader application domains (e.g., law, medicine), and an "attack" represents either a model failure or an adversarial prompt that exploits tasks assigned to less capable models. The defender's strategy determines which model or model team to deploy for each task while anticipating that failures are most likely to occur in less-protected areas. This SSG-based approach enables incentive-aware deployment strategies that balance safety, performance, and cost while remaining robust to both adversarial threats and task-specific uncertainties in mission-critical settings.

\noindent\textbf{Key challenge(s).}
SSG-based deployment plans must operate in real time within practical, large-scale systems, which may require moving beyond single-stage formulations toward multi-stage models that capture evolving deployment dynamics. A major challenge is that the true risk profiles of different models across tasks are often unknown and must be learned online. Unlike traditional SSG settings with well-defined, discrete targets, LLM deployment involves a vast and open-ended task space where failure modes can be subtle, delayed, or emerge only through model interactions, causing the underlying payoff structure to shift rapidly. Designing SSG frameworks that can learn these payoffs while simultaneously optimizing deployment strategies remains a key challenge.

\section{Conclusion}
This paper suggests that AI safety can benefit from viewing oversight as a strategic, resource-constrained problem rather than a purely model-centric one.
Stackelberg Security Games provide a principled lens for this shift, offering a way to reason about limited auditing capacity, adversarial uncertainty, and the multi-stage structure of modern AI pipelines. Our three directions in the lifecycle of LLMs illustrate how SSGs can unify disparate safety challenges under a single strategic framework.
Looking forward, this perspective opens a broader research agenda: developing richer models of adversarial behavior, designing scalable oversight algorithms, and building empirical testbeds that capture the dynamics of real-world AI systems. By embracing SSGs, AI safety can move toward more anticipatory and resilient forms of oversight capable of withstanding the strategic pressures that will accompany increasingly capable AI systems.

%%%%%%%%%%%%%%%%%%%%%%%%%%%%%%%%%%%%%%%%%%%%%%%%%%%%%%%%%%%%%%%%%%%%%%%%

%%% The acknowledgments section is defined using the "acks" environment
%%% (rather than an unnumbered section). The use of this environment 
%%% ensures the proper identification of the section in the article 
%%% metadata as well as the consistent spelling of the heading.

\begin{acks}
This work was supported by ONR MURI N00014-24-1-2742.
\end{acks}

%%%%%%%%%%%%%%%%%%%%%%%%%%%%%%%%%%%%%%%%%%%%%%%%%%%%%%%%%%%%%%%%%%%%%%%%

%%% The next two lines define, first, the bibliography style to be 
%%% applied, and, second, the bibliography file to be used.

\bibliographystyle{ACM-Reference-Format} 
\bibliography{sample}

%%%%%%%%%%%%%%%%%%%%%%%%%%%%%%%%%%%%%%%%%%%%%%%%%%%%%%%%%%%%%%%%%%%%%%%%

\end{document}